\documentclass[11pt,a4paper]{article}
\usepackage[hyperref]{acl2020}
\usepackage{times}
\usepackage{latexsym}
\usepackage{graphicx}

\usepackage{microtype}
\usepackage{hyperref}

\aclfinalcopy

\title{Novel Building Detection and Location Intelligence in Aerial Satellite Imagery}

\author{Sandeep Singh \\
  \texttt{ssingh600@gatech.edu} \\\And
  Christian Wiles \\
  \texttt{cwiles7@gatech.edu} \\\And
  Ahmed Bilal \\
  \texttt{abilal7@gatech.edu} \\
  }

\date{}

\begin{document}
\maketitle
\begin{abstract}
Building structures detection and information about these buildings in aerial images is an important solution for city planning and management, land use analysis. It can be the center-piece to answer important questions such as planning evacuation routes in case of an earthquake, flood management, etc. These applications rely on being able to accurately retrieve up-to-date information. Being able to accurately detect buildings in a bounding box centered on a specific latitude-longitude value can help greatly. The key challenge is to be able to detect buildings which can be commercial, industrial, hut settlements, or skyscrapers. Once we are able to detect such buildings, our goal will be to cluster and categorize similar types of buildings together.

\end{abstract}

\section{Introduction}
We plan on reproducing semantic segmentation CNN models based on U-Net \cite{ronneberger2015unet} and Res-U-net \cite{Diakogiannis_2020} algorithms, trained by transfer learning using the ImageNet dataset. In addition, we will optimize these different models with focal loss, dice loss, cross entropy loss, hierarchical loss and differently weighted intersection-over-union (IoU) loss to overcome issues of scale difference [3] in building detection. To clearly delineate each individual member's contribution, we will be organizing our paper by each member's contribution.

\section{Methodology}
We have decided to work on three different models in parallel, which are inherently different on following parameters:
\begin{enumerate}
\item Architecture of models: We have decided to use different variations of UNet architectures, which might help us focus of different patterns easily.
\item Pre-training usage of the models: We have employed training from scratch as well using pre-trained weight for the encoder layers across out 3 of the models. 
\item Loss functions being employed by models: We have used different loss functions in all the different models. 
\item Image sampling and augmentations employed: We have used totally different techniques to samples the images from the data sets available to introduce the component of randomness in data distribution across models. While model 1 and 2 have used random resized crops of size 224x224 from big 5000x5000 images and masks; model 3 has resorted to fixed size split of big 5000x5000 images in 512x512 tiles. Both these sampling techniques have again used totally different set of transforms with different quantifiers. 
\end{enumerate}
While we have details of all the models provided in the sections 3,4 and 5, here is a quick overview of the ensemble technique exploited by us: 
\begin{enumerate}
\item Get prediction from all 3 models. All predictions must of same size as input image. We output thresholded mask as tensor of softmax probabilities. 
\item Chose the most confident pixel from each of these masks from 3 models for every pixel location's softmax probability. This is will be single merged mask of same size as input. 
\item From merged mask predicted, drop anything that is not at least 0.75 confident. Rational behind doing this is that we have already chose most confident pixel locations from 3 models masks, So, all the softmax value at this stage must be pretty confident ones at least in one of the models. 
\end{enumerate}

As seen in the fig 1. We have final mask, which is without any less confident pixel.
\begin{figure*}[h]
    \centering
    \includegraphics[width=\textwidth]{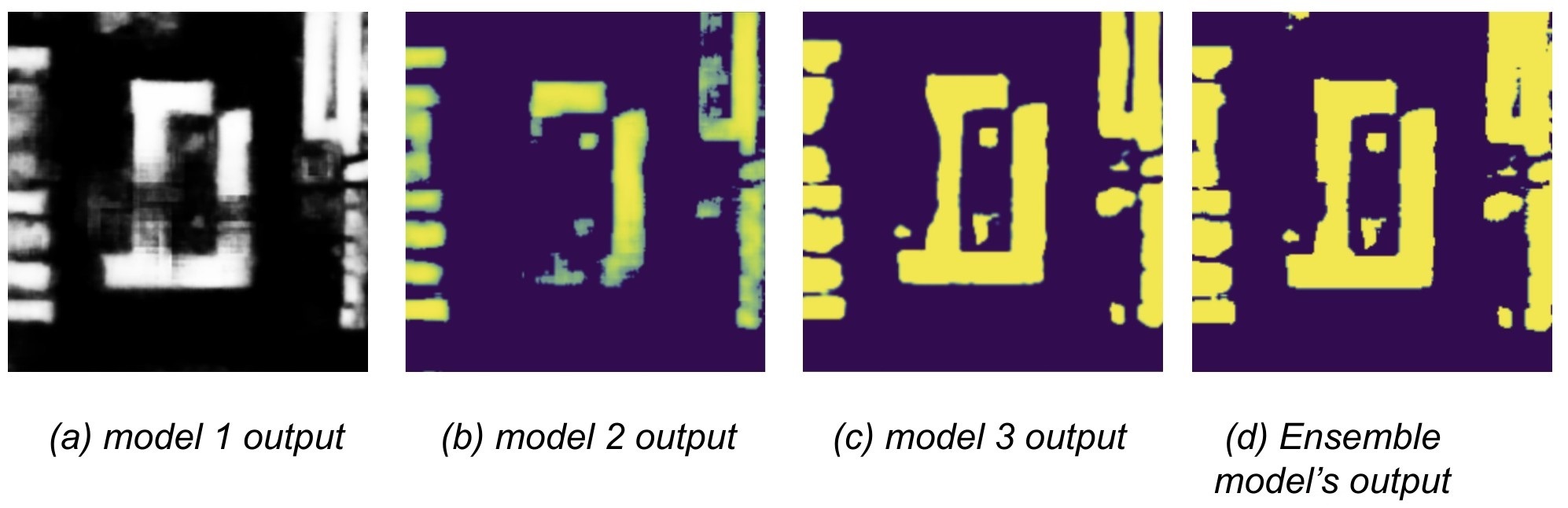}
    \caption{Ensemble method being employed. Output softmax probabilities are compared and chosen to select most confident pixels of 2 models and thresholded to be at least 75\% confident to be considered for final mask.}
    \label{fig:ensemble}
\end{figure*}

\section{Standard U-Net Architecture}
As all three models are derived from U-net, we will begin with a discussion of the canonical U-net architecture. U-net is a modular model largely consisted of encoder blocks and decoder blocks. A graphical depiction of a u-net model can be found in Figure \ref{fig:u_net_arch}.
\subsection{Encoder blocks}
The encoder layer is primarily responsible for detecting the 'what' elements of the images. The goal is to be able to extract features in the image at different scales and different levels of abstraction. As such, at every steps of the encoder, two 2D convolutional blocks are used to extract information from the image and double the size of the feature space. At each encoder layer, we used a maxpool layers of 2x2 kernel size and a stride of 2 for down sampling spatially. This allowed us to increase the number of filters at each of our encoder layers without being extremely computationally expensive and increase the receptive field of our filters with deeper layers allowing for segment detection at multiple scales.

\subsection{Decoder blocks}
The decoder layers are the up sampling layers in the model. The primary purpose of these layers is to localize the features extracted in the encoder block. This information is essential in our semantic segmentation in order to be able to output an image with the buildings detected localized in the right spaces in our output mask. For up sampling, we used Transposed Convolution layers. This allows us to ultimately assign class labels to each pixel in our image as part of our semantic segmentation.\\
At each of our decoder layers, we also make use of skipped connection given to us by the respective encoder layer for our decoder layer. The skipped connections cross from same sized part in the encoders to the decoders. The skipped connections allow us to overcome problem of vanishing gradient, increasing dimensionality and help regain the initial spatial information that we lost during the encoding path. 

\subsection{Integration}
A full u-net model is composed of N encoder blocks, and N-1 decoder blocks. The feature space of the first encoder block is a hyperparameter but seems to be often set to 64 or 128. Save for the last encoder block, the output of the final convolutional layer in each encoder block is cropped and concatenated with the output of the transposed convolutional layer of the decoder block. At the end of the model, a 1x1 convolutional layer is used to create a classifier head with the same features space as the number of classes. This can be passed to a softmax layer to produce class probabilities.

\section{Model 1 approach (Christian)}
\subsection{Data Pipeline and Exploration}
For this project, we used the Inria Aerial Image Labelling Dataset for training \cite{maggiori2017dataset}. The dataset consists of 360 (180 train and 180 test) 5000x5000 pixel full-color images with corresponding masks indicating the presence of building or non-building pixels. Only the training set had ground-truth segmentation masks available. Images were taken from a variety of settings, including rural and urban cities from different continents. A few problems had to be solved to enable training with this dataset: data augmentation and quick random access.
\subsubsection{Data pre-processing and augmentation}
To generate more data for training, data augmentation was undertaken. Pytorch's standard transformation library does not make allowances for maintaining consistent transformations between an image and a segmentation mask, so the functional transforms library was used, which allows for the randomness to be provided by external variables, which can be held static between the ground truth and full-color images.
For each of the 180 input training images, 350 224x224x3 image patches were created. This was intended to allow transfer learning for networks trained on ImageNet. A patch was taken from the image with random width (between 100 and 500 px), height (within +/- 10\% of the random width), and image origin. This image was then randomly flipped horizontally and/or vertically and normalized by ImageNet standard deviation and mean. The intention of this was to teach the model scale and orientation invariant features.
Once the incoming data was processed, it was split 80/20 into train and validation sets. Data from all 5 cities in the dataset was randomly selected for both validation and training set, as the test set consists of different cities and would be usable for testing how well the model can generalize. A manual seed was set such that this split was repeatable if restarting training from a checkpoint.
\subsubsection{Caching}
It was found that performing the loading of the full-color images and performing transformations on the fly was too computationally intensive. To alleviate this problem, after the first time the transformations were done, the resulting input and target tensors were saved to disk. This reduced the time to train significantly, as a 70 MB image did not have to be loaded and manipulated thousands of times per epoch, but instead a 600 KB tensor could be used.
\subsubsection{Other Considerations}
It was also deemed important to add support for visualization of training-related metrics. Tensorboard support was added to the project to track training and validation set loss and accuracy, precision-recall curves for the validation set, and visualization of the forward pass of the model on the validation set. One last consideration made was the use of a seed when splitting the training and validation set. It was noticed that when resuming training from a checkpoint that the validation set was not the same as before the checkpoint. By maintaining the constant seed, a barrier was maintained between the two sets.
\subsubsection{Dataset Statistics}
As is often the case with segmentation tasks, the dataset was not balanced between building pixels and non-building pixels. The training dataset was analyzed to determine the prevelance of each class. The findings are below.

\subsection{U-net from scratch in Pytorch}

To test the hypothesis that building detection was a sufficiently specific domain to merit training from scratch, a u-net was created with random Xavier-initialized weights. 4 encoder and 3 decoder blocks were used, with the first encoder block having a hidden dimension of 64 features. Both the original U-net paper \cite{ronneberger2015unet} and Johannes Schmidt's blog posts \cite{schmidt_2021} were consulted in the creation of the model. 

Two major deviations were attempted from the models mentioned above. Both u-nets resulted in a cropped image with every convolution due to the use of valid padding. By using same padding, on convolutions and transposed convolutions, we can return an image that is of equal size to the input. This may result in slightly worse accuracy in the extremities of the image due to the extrapolation employed by same padding, but does simplify some aspects of the analysis, as every mask pixel has a corresponding prediction.

Secondly as the output of a 2-class softmax classifier only has 1 degree of freedom, it was attempted to perform classification as a single-class regression, with the output of the regression put through a sigmoid function. This one-channel output can then be interpreted as p(building). This approach was eventually discarded, as it had a very small impact on model size due to only affecting the final 1x1 convolutional layer, and adding a second classifier dimension increased model performance by a few percentage points.

Finally, batch normalization was added between the convolutional and activation layers in encoder and decoder blocks. These recenter the distribution of the output of the convolutional layers and add to stability in training as seen in \cite{santurkar2019does}. 

To discourage overfitting, dropout was added between the output of the final decoder layer and the 1x1 convolution. 

\subsubsection{Loss function}
3 different loss funcitons were attempted with this model. With the regression-based approach, weighted mean square error was used due to the class imbalance. On a per-batch basis, the effective number of building and non-building pixels was calculated on a per-batch basis, similar to the methodology in \cite{cui2019classbalanced}. This weighting was weigh the loss on the minority class (buildings) more heavily.

Once the model had progressed to a two-class method, two losses were pursued: weighted cross-entropy loss and dice loss. Dice loss was pursued due to its background in segmentation tasks and invariance with respect to class imbalance (as dice loss is related to the size of the true positive region). Binary cross-entropy weighted by the inverse of effective number was also explored. This provides a more convex loss function that should be easier and more stable to train.

\subsubsection{Training}
Training was performed over 20 epochs with the Adam optimizer. No learning rate scheduler was used to govern learning rate as epochs progressed, as Adam should manage its own learning rates on a per-parameter basis \cite{kingma2017adam}. A batch size of 20 was used to fit in GPU memory. The model state was saved to disk whenever validation accuracy exceeded the previous maximum to allow for training to be resumed later.

\subsubsection{Results}

\begin{table}[h!]
\centering
\begin{tabular}{|c|c|c|c|c|} 
 \hline
 Loss Fn & Accuracy & IOU Score & F1-Score \\ [0.5ex] 
 \hline
 Dice & 94.9\% & 0.717 & 0.836 \\ 
 \hline
 BCE & 95.0\% & 0.726 & 0.841 \\ 
 \hline
\end{tabular}
\caption{Model 1 Validation Set results after 20 Epochs}
\label{table:model_1_results}
\end{table}

\begin{figure}[h]
    \centering
    \includegraphics[width=0.45\textwidth]{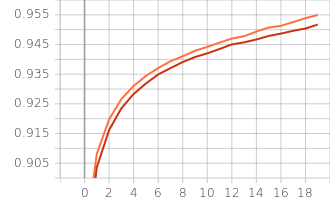}
    \caption{Training accuracy over 20 epochs for model 1. Weighted CE in orange, Dice Loss in red.}
    \label{fig:train_acc}
\end{figure}

\begin{figure}[h]
    \centering
    \includegraphics[width=0.45\textwidth]{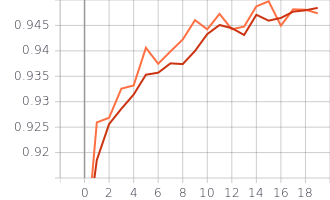}
    \caption{Validation accuracy over 20 epochs for model 1. Weighted CE in orange, Dice Loss in red.}
    \label{fig:val_acc}
\end{figure}

Weighted cross-entropy resulted in marginally better training efficiency and overall metrics, but by an almost negligible amount. For both approaches, overfitting does not appear to be a concern, as the validation and training accuracy are almost identical. 

Though this model performed worse than Model 2, it is hard to say whether it is due to differences in pre-training or the squeeze-and-attention layers. It is likely, however, that due to learning features from scratch, it may provide diversity in the ensemble that can help in overall accuracy.

\section{Model 2 approach (Ahmed)}

\subsection{Model Specifications}
\subsubsection{Double Convolution Blocks}
The Unet build consisted of a double convolution layer, where each convolution layer consisted of a kernel size of 3, stride and padding of 1.  We set the bias to false in order to add a BatchNorm layer, which is then followed by a ReLU activation layer. We settled on a small 3X3 kernel receptive field in our convolution layers in order to be able to detect very small edges and shapes in our aerial images. Doing so is especially relevant for our aerial images as there is a lot of noise in the images and our model needs to be able to use small edges and shapes to detect buildings as buildings appear in many different sizes in our input images.\\
\subsubsection{Encoder Layers}
The authors in \textit{U-Net: Convolutional Networks for Biomedical
Image Segmentation} \cite{ronneberger2015unet} recommend encoding layers with output channels 64,128,256, 512 and 1024. However, we found more success with output channel layers 16,32,64,128 and 256. We believe this is because lower output channels of 16 and 32 in the start allow us to detect really small building segments with a small receptive field. In addition, the 512 and 1024 channel layers were not leading to any significant performance gains in our testing.

\subsubsection{Decoder Layers}
At each of our decoder layers, we also make use of skipped connection given to us by the respective encoder layer for our decoder layer. The skipped connections cross from same sized part in the encoders to the decoders. The skipped connections allow us to overcome problem of vanishing gradient, increasing dimensionality and help regain the initial spatial information that we lost during the encoding path. 

\subsection{Pre-trained ResNet34 Encoder Specifications}
We now add ResNet 34 Encoder layers to the model. As such, we are now performing the Double Convolution blocks 3,4,6, and 3 times at each encoder layer level, using skipped connections between encoder layers, and using a higher stride to down sample instead of max pooling. These encoder layers are also pre-trained on image-net dataset. \\
\subsubsection{Encoder Modifications}
After finding success in our U-Net built with 16,32 and 64 output channel initial encoder layers, we replace the initial ResNet convolution, ReLU and Max Pool layers with our U-Net 16,32 and 64 output channel encoder layers with skipped connections in order to preserve a lot of the small shapes and edges information in our images.
\subsubsection{Attention Mechanism}
For the loss function, we will be using the Dice Loss to create cleaner mask segments to represent the buildings. In order to supplement our model in reducing the Dice Loss, we also include an attention mechanism using spatial and channel 'squeeze \& excitation' Blocks. This is done to aid our encoder layers in spatial encoding for more accurate mask prediction and better network flow. The authors in \textit{Recalibrating Fully Convolutional Networks with Spatial and Channel 'Squeeze \& Excitation' Blocks} \cite{DBLP:journals/corr/abs-1808-08127} found a reduction of 4-9\%  in the Dice Loss. We see similar results in our testing.
\subsubsection{Results}
In our testing, we saw the pre-trained image-net backbone significantly increase the model performance. After 15 epochs, we saw the following results.\\
\begin{table}[h!]
\centering
\begin{tabular}{|c|c c c c|} 
 \hline
 Set & Acc & Loss & IOU Score & F1-Score \\ [0.5ex] 
 \hline
 Train & 0.965 & 0.116 & 0.749 & 0.856 \\ 
 \hline
 Val & 0.961 & 0.129 & 0.702 & 0.824 \\ 
 \hline
\end{tabular}
\caption{Validation Set results after 15 Epochs}
\label{table:1}
\end{table}\\
Detailed Model 2 results and mask outputs available in appendix Section F

\section{Model 3 Distinctive Approach (Sandeep)}
\subsection{Data Sampling Strategy}
For this model, we have used ``Progressive Resizing.'' This is a training technique where we purposefully change the contents of image by resizing the images to contain more area. Instead of randomized crops of size 224x224, we created 512x512 non-overlapping and contiguous tiles, only then resizing them to 224x224 input images. On average, each tile has almost 4 times more buildings in each tile compared to model 1 and 2. Hence, the model more easily learns smaller buildings in more crowded areas \cite{progressiveResizing}.

\subsection{Architectural Considerations} For model 3, we have evaluated 3 types of encoders (Resnet18/34/50) and choose ResNet34 as ResNet18 has shown to be struggling to encode features of smaller buildings successfully.  ResNet34 and 50 have shown very similar, but resnet 50 slow performance in detected buildings without any marginal increase in performance. One more significant improvement in model 3 was the use of we used Pixel Shuffle up-sampling in the decoder blocks, as provided by shuffleblock implementation \cite{pixelshuffle}.

\subsection{Training and Validation Split} For model 3, we have done the data split on the basis of geography, instead of random ratio split. Out of  training data from 5 cities as: Austin, Chicago, Kitsap, Tyrol, Vienna. Different cities are included in each of the subsets. e.g., images over Chicago are included in the training set, but not on the test set. Also, images over San Francisco are included on the test set but not on the training set.  At the same time, we have tried to include the training data all type of structures of building. e.g. low rise vs high rise vs community living buildings apartment complexes. 

\subsection{Custom Loss Function Design} Model 3 did not used CrossEntropy loss. Instead, we have written our own custom loss function, which has helped us predict foreground pixel with higher softmax confidence. We implemented Combined Loss of Dice Loss and Focal Loss with equal weights. Focal loss penalizing more confident wrong predictions more heavily. We have used gamma value of focal loss as 2. Also, Dice score has provided feed back to strive to keep precision and recall both highest possible. Also, for model 3, we have used the Dice Score metric.

\subsection{Training Convergence} For model 3, we used a LR finder scheduler before starting to fine tune the pretrained weights of ResNet34 encoder. This has helped of find the most appropriate maximum LR value. The second innovative technique employed by us was "Fit-one-cycle" \cite{fitonecycle} to achieve super convergence. In this technique, we increase LR to maximum value in initial batches before start to anneal the learning rates. Please refer to figure for LR finder and fit one cycle both \cite{AdamWandSuperConvergence}. This technique is taken from Leslie Smith iconic Super Convergence paper. 
Also, We trained with total 40 epochs without over-fitting and saved the model only when better score on validation was seen without over-fitting, while Dice score was approaching 92\%. 

\begin{table}[h!]
\centering
\begin{tabular}{||c c c||} 
 \hline
 Train\_Loss & Valid\_Loss &  Dice\_Score \\  
 \hline
 0.102414 & 0.115400 & 0.920870\\
 \hline
\end{tabular}
\caption{ Model 3 Losses and Metrics Values}
\label{table:5}
\end{table}

\begin{figure}
    \centering
    \includegraphics[width=0.45\textwidth]{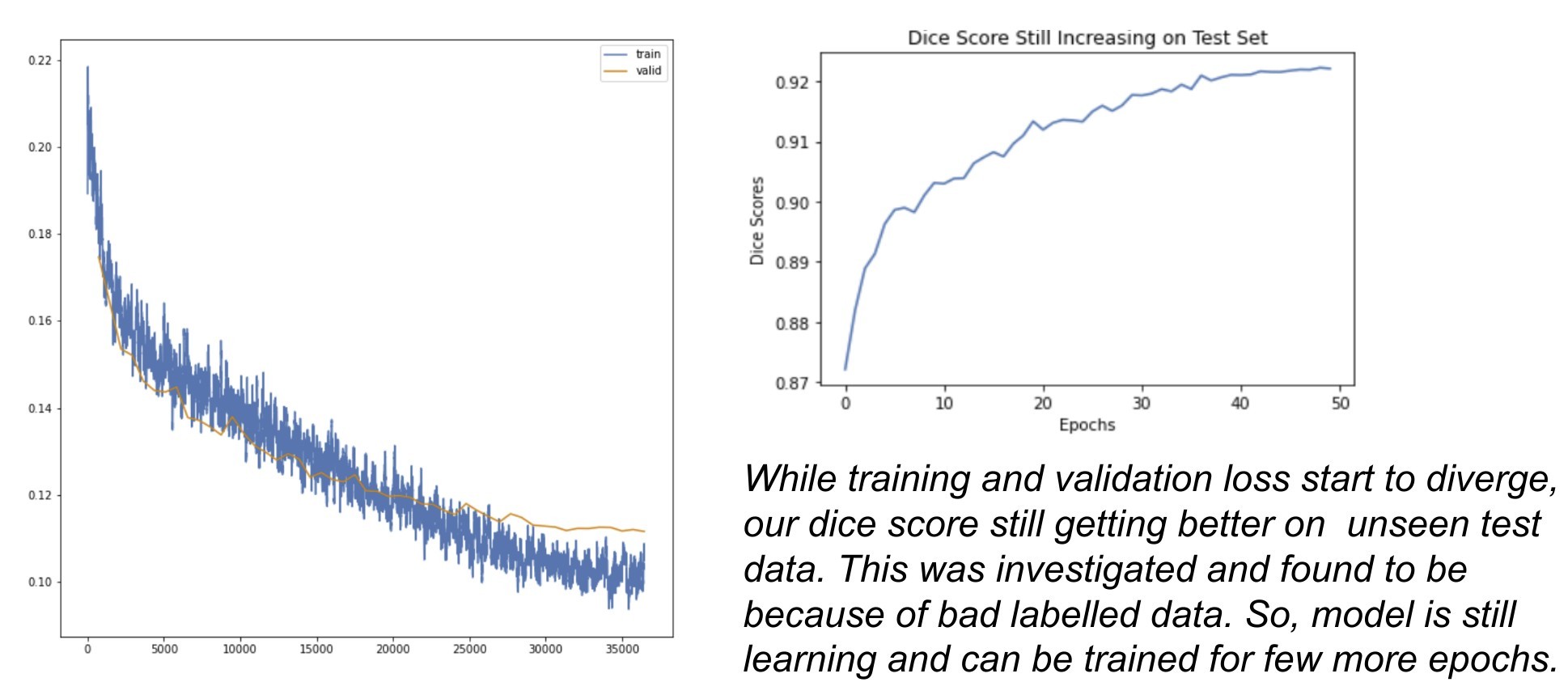}
    \caption{Model 3 Training Curves.}
    \label{fig:model3curves}
\end{figure}

\subsection{Post Processing Enhancements} As can be seen in some our validation images, the masks created by this model are not always crisp and polygonal. Please refer to appendix figure for Model 3's traditional computer vision techniques of segmentation and attempt to fuse the results with Unet predicted mask. We have tried to experiment with Otsu's threshold, Watershed segmentation, SLIC Super pixel algorithm for segmentation. We have selected all the coincidental mask segments from super pixel algorithms' output with Unet's mask and tried to shape correct polygon for buildings with sharp edges (see fig. \ref{fig:post_process}).

\section{Experiences and Challenges}

\subsection{Challenges}
Satellite images are very noisy and affected by many factors such as weather, zoom level, resolution, trees, and cost to obtain \cite{satelliteimages}. After evaluating, we used Google Static maps API for pulling additional data because of quality and ease of use.
 
\subsection{Project Success Criterion}
 We were able to build model with Dice score more than 90 percent on test set and model 2 achieving detection of more 96 percent ground truth pixels in valid set. With help of Softmax based adaptive selection and ensemble, we have achieved detection of more 93 percent ground truth pixels in test set.
 
\subsection{Conclusion and Future Aspirations}
We have explored and confirmed that deep learning based segmentation is very effective in segmenting buildings. We could extend these models for clustering similar buildings, classifying residential vs commercial, predicting future constructions or detecting illegal construction or activities. Potential technical improvements include learned polygonization, elastic transformations during training, and model compression for cheaper inference.

\clearpage
\bibliographystyle{acl_natbib}
\bibliography{acl2020.bib}

\begin{thebibliography}{13}
\expandafter\ifx\csname natexlab\endcsname\relax\def\natexlab#1{#1}\fi

\bibitem[{Aitken et~al.(2017)Aitken, Ledig, Theis, Caballero, Wang, and
  Shi}]{pixelshuffle}
Andrew Aitken, Christian Ledig, Lucas Theis, Jose Caballero, Zehan Wang, and
  Wenzhe Shi. 2017.
\newblock \href {http://arxiv.org/abs/1707.02937} {Checkerboard artifact free
  sub-pixel convolution: A note on sub-pixel convolution, resize convolution
  and convolution resize}.

\bibitem[{Cui et~al.(2019)Cui, Jia, Lin, Song, and
  Belongie}]{cui2019classbalanced}
Yin Cui, Menglin Jia, Tsung-Yi Lin, Yang Song, and Serge Belongie. 2019.
\newblock \href {http://arxiv.org/abs/1901.05555} {Class-balanced loss based on
  effective number of samples}.

\bibitem[{Diakogiannis et~al.(2020)Diakogiannis, Waldner, Caccetta, and
  Wu}]{Diakogiannis_2020}
Foivos~I. Diakogiannis, François Waldner, Peter Caccetta, and Chen Wu. 2020.
\newblock \href {https://doi.org/10.1016/j.isprsjprs.2020.01.013} {Resunet-a: A
  deep learning framework for semantic segmentation of remotely sensed data}.
\newblock \emph{ISPRS Journal of Photogrammetry and Remote Sensing},
  162:94–114.

\bibitem[{Howard(2018)}]{progressiveResizing}
Jeremy Howard. 2018.
\newblock \href {http://arxiv.org/abs/https://fast.ai/} {Progressive resizing
  for better generalization of the computer vision models}.

\bibitem[{Kingma and Ba(2017)}]{kingma2017adam}
Diederik~P. Kingma and Jimmy Ba. 2017.
\newblock \href {http://arxiv.org/abs/1412.6980} {Adam: A method for stochastic
  optimization}.

\bibitem[{Maggiori et~al.(2017)Maggiori, Tarabalka, Charpiat, and
  Alliez}]{maggiori2017dataset}
Emmanuel Maggiori, Yuliya Tarabalka, Guillaume Charpiat, and Pierre Alliez.
  2017.
\newblock Can semantic labeling methods generalize to any city? the inria
  aerial image labeling benchmark.
\newblock In \emph{IEEE International Geoscience and Remote Sensing Symposium
  (IGARSS)}. IEEE.

\bibitem[{Ronneberger et~al.(2015)Ronneberger, Fischer, and
  Brox}]{ronneberger2015unet}
Olaf Ronneberger, Philipp Fischer, and Thomas Brox. 2015.
\newblock \href {http://arxiv.org/abs/1505.04597} {U-net: Convolutional
  networks for biomedical image segmentation}.

\bibitem[{Roy et~al.(2018)Roy, Navab, and
  Wachinger}]{DBLP:journals/corr/abs-1808-08127}
Abhijit~Guha Roy, Nassir Navab, and Christian Wachinger. 2018.
\newblock Recalibrating fully convolutional networks with spatial and channel
  'squeeze {\&} excitation' blocks.
\newblock \emph{CoRR}, abs/1808.08127.

\bibitem[{Santurkar et~al.(2019)Santurkar, Tsipras, Ilyas, and
  Madry}]{santurkar2019does}
Shibani Santurkar, Dimitris Tsipras, Andrew Ilyas, and Aleksander Madry. 2019.
\newblock \href {http://arxiv.org/abs/1805.11604} {How does batch normalization
  help optimization?}

\bibitem[{Schmidt(2021)}]{schmidt_2021}
Johannes Schmidt. 2021.
\newblock \href
  {https://towardsdatascience.com/creating-and-training-a-u-net-model-with-pytorch-for-2d-3d-semantic-segmentation-dataset-fb1f7f80fe55}
  {Creating and training a u-net model with pytorch for 2d \& 3d semantic
  segmentation: Dataset...}

\bibitem[{Smith(2018)}]{fitonecycle}
Leslie~N. Smith. 2018.
\newblock \href {http://arxiv.org/abs/1803.09820v2} {A disciplined approach to
  neural network hyper-parameters: Part 1 – learning rate,batch size,
  momentum, and weight decay}.

\bibitem[{Sylvain~Gugger(2018)}]{AdamWandSuperConvergence}
Jeremy~Howard Sylvain~Gugger. 2018.
\newblock \href
  {http://arxiv.org/abs/https://www.fast.ai/2018/07/02/adam-weight-decay/}
  {Adamw and super-convergence is now the fastest way to train neural nets}.

\bibitem[{Wikipedia(2018)}]{satelliteimages}
Wikipedia. 2018.
\newblock \href
  {http://arxiv.org/abs/https://en.wikipedia.org/wiki/Satellite_imagery}
  {Satellite imagery and challenges associated with them}.

\end{thebibliography}

\appendix

\section{Code Repository}
We have work on total of 4 repositories during our project life cycle. 3 repositories where used by each of us individually and 1 final repository was created as a place to perform ensemble and integration of all models together and do all the needed post-processing. Here the repositories as below:
\begin{itemize}
\item Integrated Final repo:\\
\url{https://github.com/sandeepsign/building\_footprint\_ensemble}
\item Ahmed's repo:\\
\url{https://github.com/abilal19/DL\_FinalProject\_Draft}
\item Christian's repo:\\
\url{https://github.com/cswksu/aerialDetection}
\item Sandeep's repo:\\
\url{https://github.com/sandeepsign/building\_footprints\_cs7643}
\end{itemize}

\section{Individual Contributions}
\begin{table}[h]
\begin{center}
\resizebox{0.45\textwidth}{!}{%
\begin{tabular}{ |c|c| }
 \hline
 Contributor & Contribution \\ 
 \hline
 Ahmed & Full Model 2 (best performing),\\ & Project Report \\ 
 \hline
 Christian & Full Model 1, data preprocessing \\ & and visualization for models \\ & 1 and 2, Project Report \\ 
 \hline
 Sandeep & Full Model 3, ensembling, \\ & post processing, Geo Coding, VM Setup,\\ &  Clustering of polygons, Project Report \\ 
 \hline
\end{tabular}}
\caption{\label{tab:contributions}Individual team member contributions.}
\end{center}
\end{table}

\section{Data Source}
We have used INRIA's spacenet challenge data from:\\
https://project.inria.fr/aerialimagelabeling/

\section{Train Infrastructure}
We have used combination of techniques to execute this project. 

Each had used individual hardware for setup and eventually to run long time training on google cloud VM instance. 

GPUs Used are:\\
nVIDIA QUADRO RTX 5000 16GB \\
nVIDIA T4 16GB\\
nVIDIA GTX 1080ti 11GB\\

\section{Enlarged Figures}

\begin{figure*}[h]
    \centering
    \includegraphics[width=\textwidth]{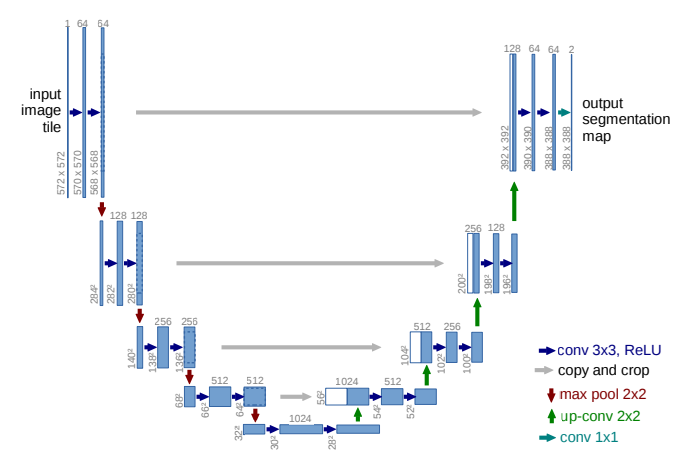}
    \caption{Example of overall organization of u-net model from \cite{ronneberger2015unet}}
    \label{fig:u_net_arch}
\end{figure*}

\begin{table}[h]
\begin{center}
\resizebox{0.4\textwidth}{!}{%
\begin{tabular}{ |c|c|c|c| }
 \hline
 Class & Number of Pixels & Percentage & Effective Number \\ & & &  ($\beta$ = 1 - 10E-9) \\ 
 \hline
 Building & 7.1E8 & 1.58\% & 5.08E8 \\ 
 Not building & 4.4E10 & 98.4\% & 1.00E9 \\ 
 Total & 4.5E10 & 100\% & 1.00E9 \\ 
 \hline
\end{tabular}}
\caption{\label{tab:pop-stats}Class distribution of pixels in training set images.}
\end{center}
\end{table}
\clearpage
\section{Model 2 Results}
\subsubsection{Model 2 Validation Set Mask Samples}
Validation Set Target\\
\includegraphics[width=0.45\textwidth]{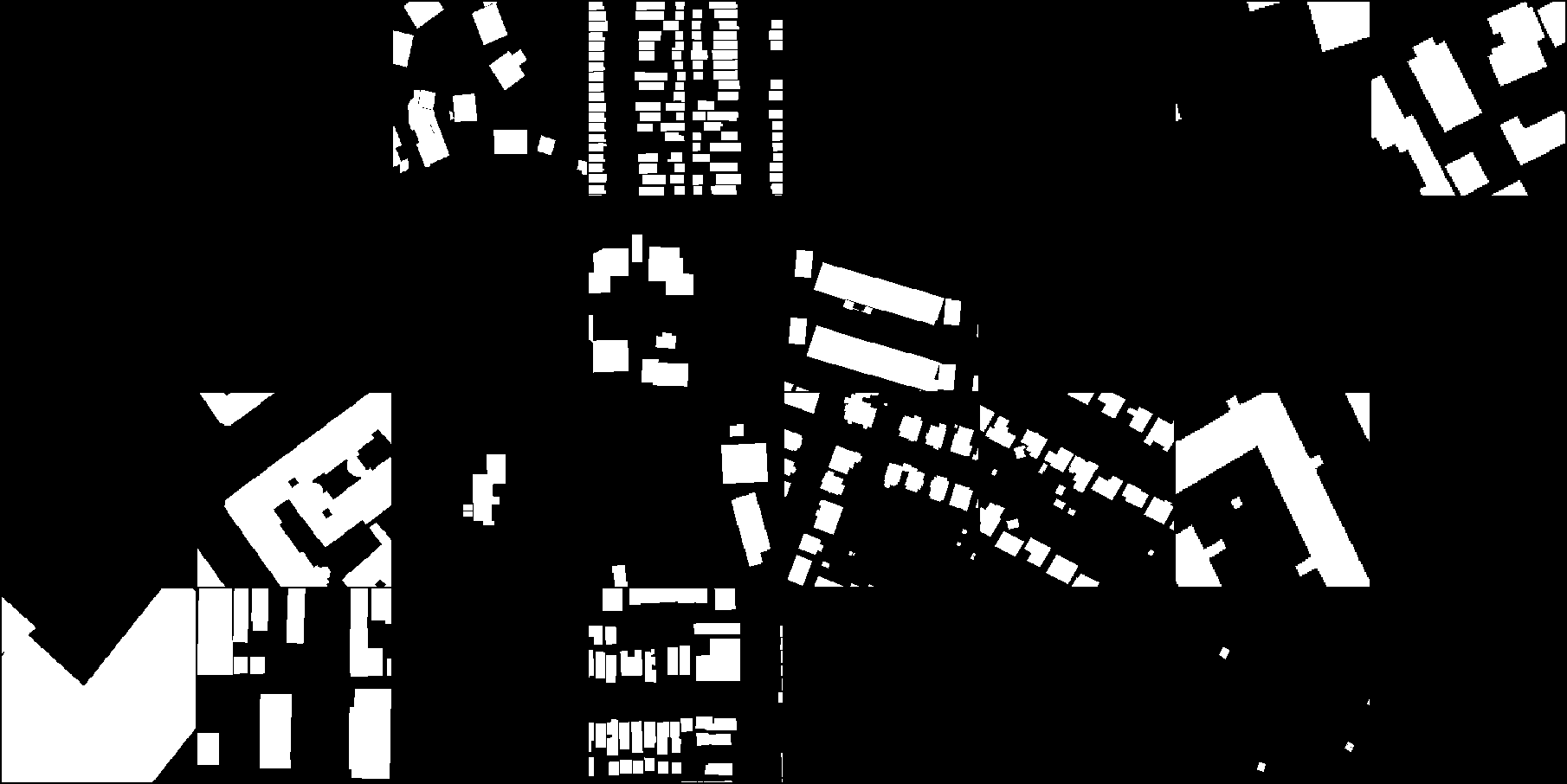}
Validation Set Output\\
\includegraphics[width=0.45\textwidth]{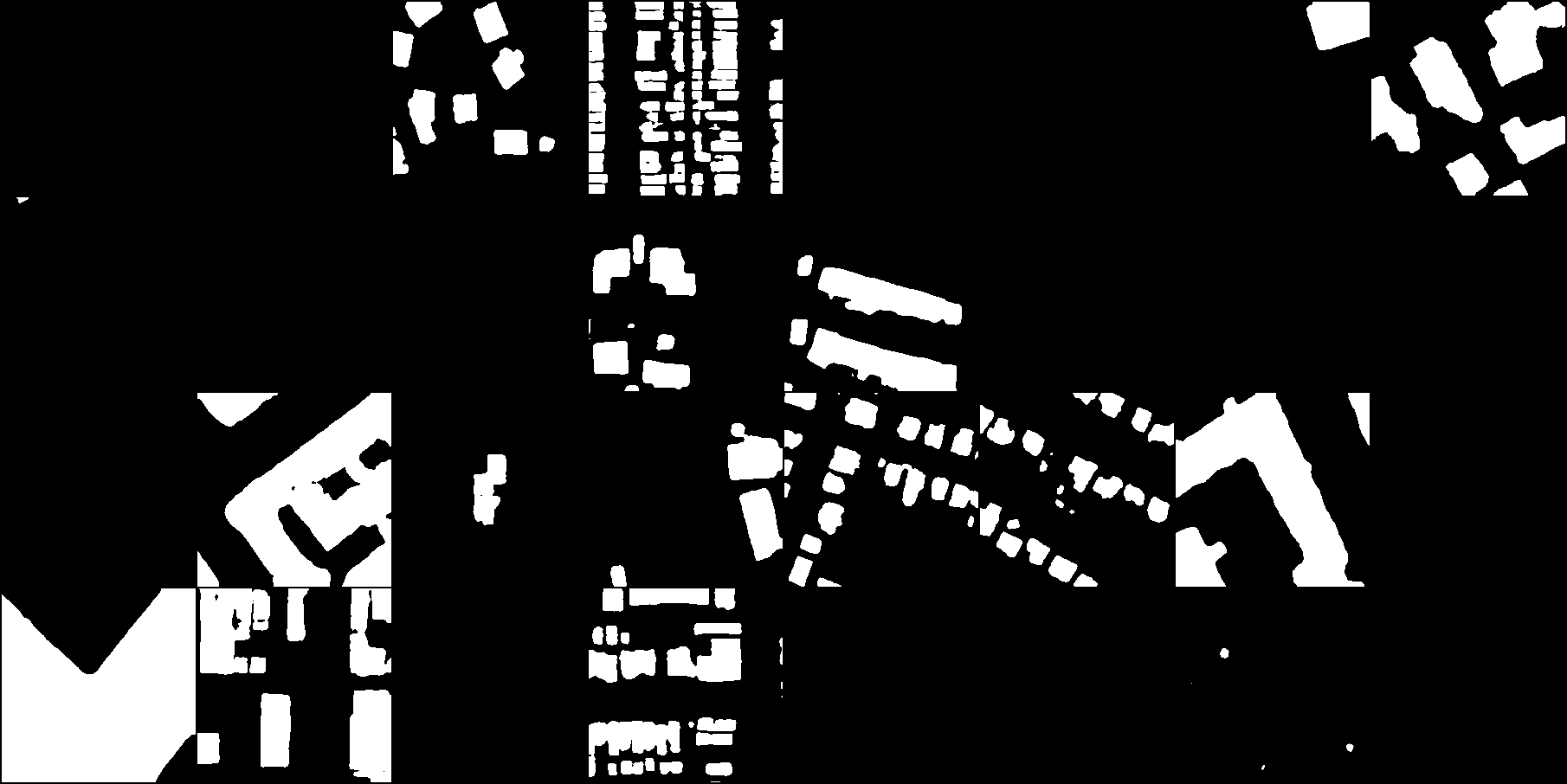}
\subsubsection{Accuracy Results Per Epoch}
\begin{figure}[h]
\includegraphics[width=0.45\textwidth]{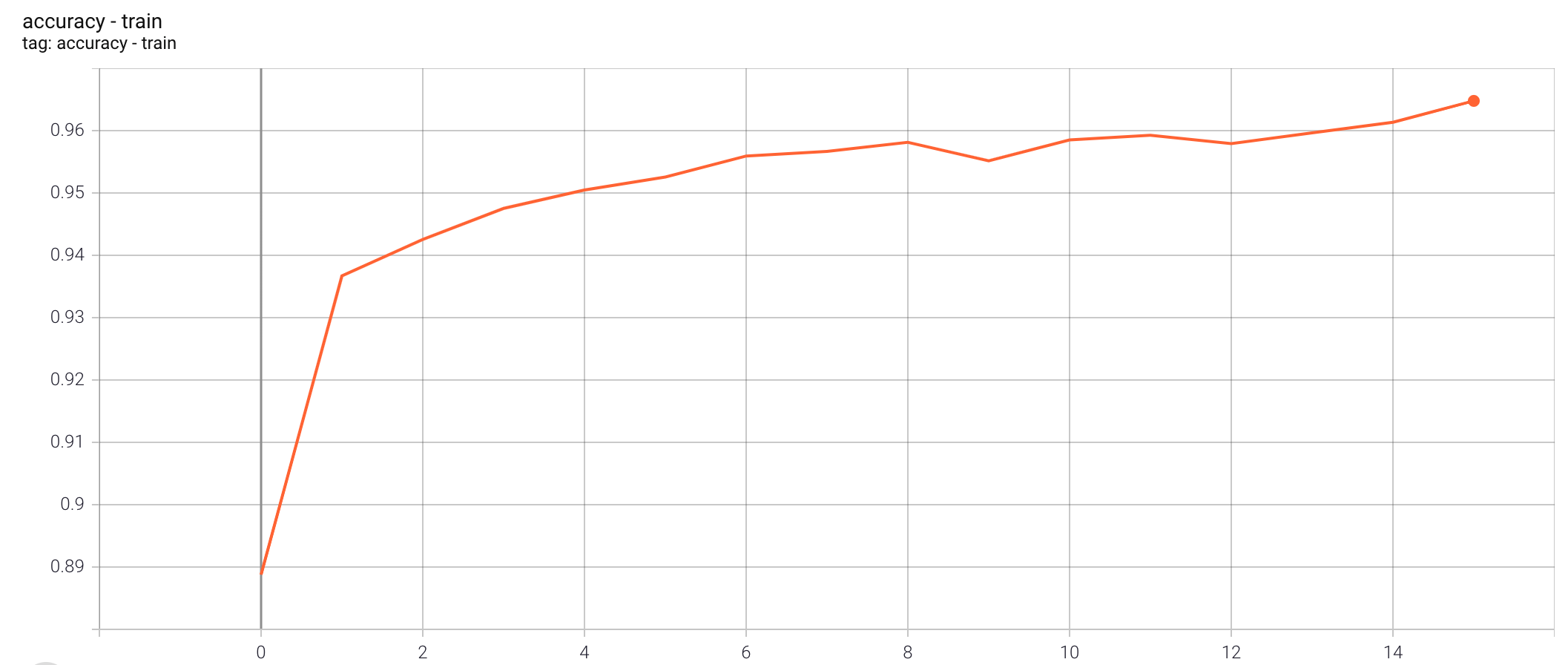}
\caption{'Model 2 Train Accuracy'}
\end{figure}
\begin{figure}[h]
\includegraphics[width=0.45\textwidth]{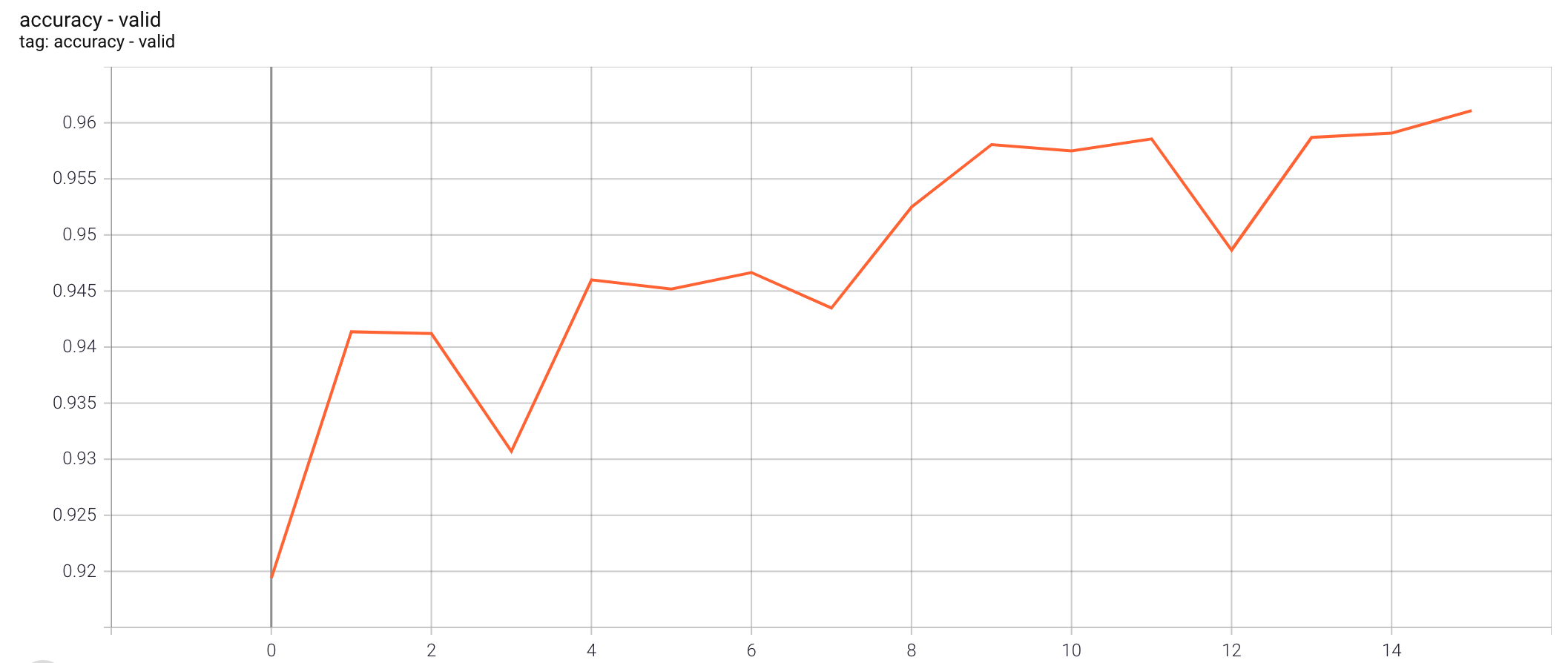}
\caption{'Model 2 Valid Accuracy'}
\end{figure}

\section{Model 3 figures}

\begin{figure}
    \centering
    \includegraphics[width=0.45\textwidth]{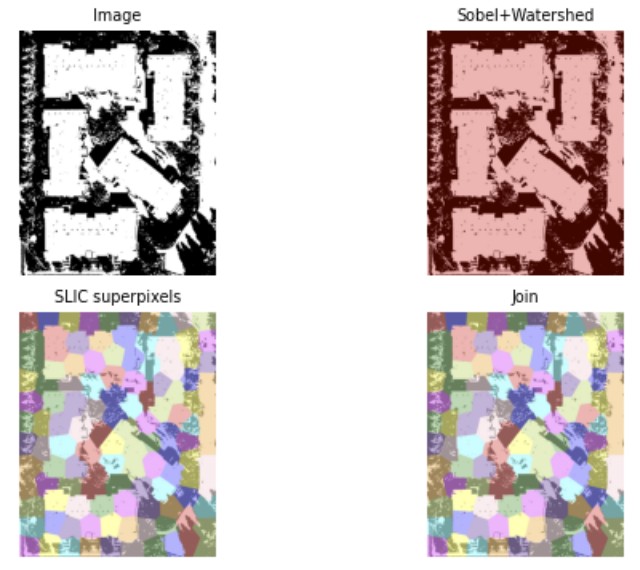}
    \caption{Post processing and merging of traditional segmentation techniques with Unet’s mask and shape correcting polygons for buildings.}
    \label{fig:post_process}
\end{figure}

\begin{figure}
    \centering
    \includegraphics[width=0.45\textwidth]{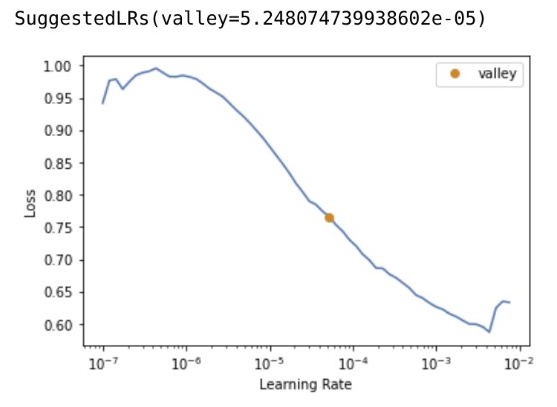}
    \label{fig:lr_finder}
    \caption{Model 3 learning rate finder}
\end{figure}

\begin{figure}
    \centering
    \includegraphics[width=0.45\textwidth]{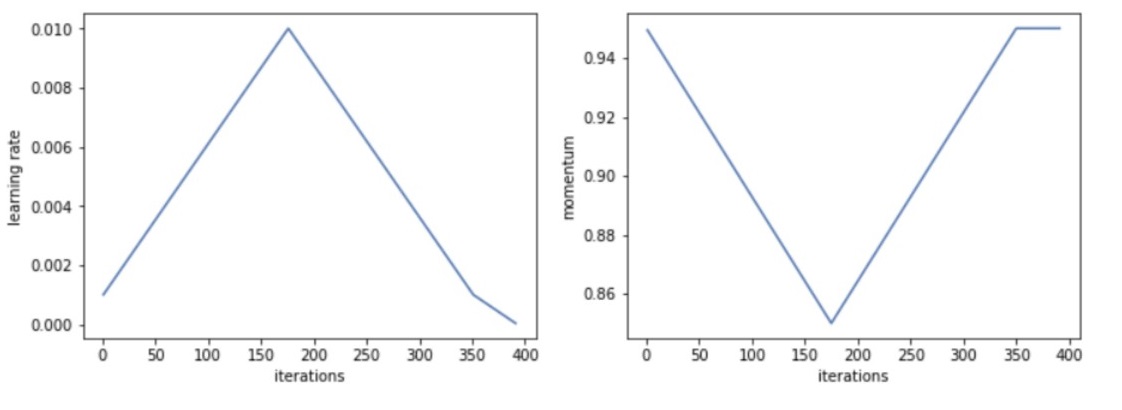}
    \label{fig:lr_scheduling}
    \caption{Model 3 fit one cycle scheduler: Momentum is moved opposite to learning rate during increase or annealing phase}
\end{figure}

\end{document}